\pdfoutput=1
\documentclass[11pt]{article}

\usepackage{acl}

\usepackage{times}
\usepackage{latexsym}

\usepackage[T1]{fontenc}
\usepackage[utf8]{inputenc}

\usepackage{microtype}

\usepackage{inconsolata}

\usepackage{enumitem}
\usepackage{naturaltableau}
\usepackage{soul}
\usepackage{booktabs}
\usepackage{linguex}
\usepackage{multirow}
\usepackage{pifont}

\colorlet{truecolor}{green!30}
\colorlet{falsecolor}{red!30}
\colorlet{rulecolor}{cyan!30}
\newcommand{\hlT}[1]{{\sethlcolor{truecolor}\hl{#1}}}
\newcommand{\hlF}[1]{{\sethlcolor{falsecolor}\hl{#1}}}
\newcommand{\surfform}[1]{\textcolor{gray}{\textsmaller[1]{\texttt{#1}}}}
\newcommand{\mlform}[1]{\textcolor{blue}{\textsmaller[1]{\texttt{#1}}}}

\newcommand{\natformT}[1]{\textcolor{black}{\textsmaller[1]{\hlT{\texttt{#1}}}}}
\newcommand{\natformF}[1]{\textcolor{black}{\textsmaller[1]{\hlF{\texttt{#1}}}}}
\newcommand{\args}[1]{\textcolor{black}{\textsmaller[1]{\hl{\texttt{#1}}}}}
\newcommand{\tokglue}{\textvisiblespace}
\newcommand{\rulecol}[1]{{\sethlcolor{rulecolor}\hl{#1}}}
\newcommand{\tabu}[2][c]{\begin{tabular}{@{}#1@{}}#2\end{tabular}}
\newcommand{\lvlno}{\textcolor{red}{\ding{55}}}
\newcommand{\lvlmaybe}{\textsmaller[2]{\textcolor{blue}{\ding{51}}}}
\newcommand{\lvlyes}{\textcolor{green!50!black}{\ding{51}}}

\renewcommand{\sectionautorefname}{\S\kern-2pt}
\renewcommand{\subsectionautorefname}{\S\kern-2pt}
\renewcommand{\subsubsectionautorefname}{\S\kern-2pt}

\title{Formal Proofs as Structured Explanations:\\
Proposing Several Tasks on Explainable Natural Language Inference}

\author{Lasha Abzianidze\\
  Institute for Language Sciences\\
  Utrecht University, Utrecht, the Netherlands\\
  \texttt{l.abzianidze@uu.nl}
  }

\begin{document}
\maketitle
\begin{abstract}
In this position paper, we propose a reasoning framework that can model the reasoning process underlying natural language inferences.
The framework is based on the semantic tableau method, a well-studied proof system in formal logic.
Like the semantic tableau, the framework is driven by refutation---something is proved if and only if its counterexample was not refuted.
Despite being rooted in formal logic, the framework shares similarities with the mental models, a theory on the psychology of reasoning.
We will show how the reasoning framework can facilitate the collection of comprehensive and structured explanations for existing naturalistic inference problems.
To make the suggestion more concrete, we propose a method of semi-automatically obtaining structured explanations from the formal proofs of a reliable and high-performing logic-based inference system. 
Taking advantage of the in-depth information available in the generated formal proofs, we show how it can be used to define natural language reasoning tasks with structured explanations.
The proposed tasks can be ordered according to difficulty defined in terms of the granularity of explanations.
We argue that the tasks that contain a natural sketch of the proofs will suffer from substantially fewer shortcomings than the existing explainable reasoning tasks (or datasets).   
\end{abstract}

%███████████████████████████████████████████████████
\section{Introduction}
\label{sec:into}

Natural Language Inference (NLI) is a popular task that evaluates NLP models on the capacity of reasoning with natural language: given a premise text and a hypothesis sentence, a model is expected to predict an inference relation (entailment, neutral, and contradiction) between the premise and the hypothesis.%
\footnote{
Originally the task was called Recognizing Textual Entailment (RTE,  \citealt{Dagan:2005}), but later the community started to refer to it as NLI mainly due to the influential works by \citet{maccartneythesis} and \citet{bowman-EtAl:2015:EMNLP}.
}  
In this way, NLI is a three-way classification task (sometimes even two-way) that is supposed to evaluate a model's capacity for drawing inferences.
To draw correct inferences, we expect a model, where possible, to interpret the relevant part of the premise and the hypothesis, employ a decision procedure that mostly follows reasonable inference rules, and use necessary linguistic semantics or knowledge (if applicable).
In contrast to the expectation, multiple works have shown that the high-performing NLI models have learned how to exploit biases in the NLI datasets rather than how to approximate sound reasoning.
The most notable example of such biases is the hypothesis-only bias \citep{poliak-etal-2018-hypothesis,gururangan-etal-2018-annotation,tsuchiya-2018-performance}, where an NLI dataset allows models to achieve a substantially high score by observing only the hypothesis sentence.%
\footnote{
There is nothing special with the hypothesis sentence in the NLI task as such.
The bias is due to the commonly used crowdsourcing method of creating a large NLI dataset where hypothesis sentences are elicited from crowd workers.
}
% Another manifestation of NLI models failing to draw inferences is insensitivity to word order    

The attempt to evaluate an entire reasoning process between a premise and a hypothesis with one of the three inference relations can be seen as a highly simplified way of modeling reasoning.
The three-way classification format of the task and unintentionally introduced biases and regularities in NLI datasets leave substantial room for deep learning-based NLI models to optimize their decision procedure in many ways, which can be meaningfully different from the sound decision procedure.
Moreover, such a setup of the task does not teach models how to provide additional information on top of the inference label, which makes it unclear why models make certain predictions.
Teaching a model to additionally generate an explanation supporting the inference label is a way to solve these two issues simultaneously: (a) to further constraint a model's decision procedure and prevent it from learning irrelevant inference relations,
(b) to provide additional insight supporting a predicted inference relation.

However, current tasks and datasets for explainable NLI or natural language reasoning suffer from several shortcomings such as:
\begin{enumerate}[label=(\Alph*),itemsep=0mm, parsep=1mm, topsep=1mm]
\item A lack of reliable automatic evaluation;
\item Coarse granularity of the explanations; 
% \item Incomplete types of explanations;
\item Incomplete set of inference relations;
\item Synthetic nature of the text and inferences.
\end{enumerate}
These shortcomings will be further discussed in \autoref{sec:shortcomings} in context with several notable explainable NLI tasks and their corresponding datasets.
Due to the severity of the shortcomings, it is not clear how they can be alleviated and how to design a more realistic and comprehensive reasoning task with reliable automatic evaluation.

The paper proposes a reasoning framework that overcomes the above shortcomings.
In particular, we introduce a framework based on the semantic tableau proof system \citep{beth:55,Hintikka:55}.
While proof systems usually operate on logical formulas that significantly differ from natural language phrases, our proof system is tailored to natural logic \citep{essays:86,maccartneythesis,Moss2010}.  
Additionally, the framework has cognitive relevance as it resembles to the theory of mental models \citep{johnsonlaird83}.  
In \autoref{sec:tasks}, we will describe the reasoning model and show how it can account for a broad range of inferences found in standard NLI datasets and, at the same time, provide several flavors of structured explanations varying in terms of comprehensiveness.
The structured nature of the explanations will facilitate reliable automatic evaluation.

%████████████
\begin{table*}[th!]
\centering
\scalebox{.95}{
\begin{tabular}{@{}l l@{}c c c c@{\kern2mm}c@{\kern2mm}c@{}}
\\\toprule
\multirow{2}{*}{Explanation type} & \multirow{2}{*}{Dataset} 
& \multirow{2}{*}{\tabu{\bf Natural-\\\bf ness}} & 
\multicolumn{3}{c}{Inf. labels}
& \multirow{2}{*}{\tabuc{\bf Comprehensive\\\bf explanation}} 
& \multirow{2}{*}{\tabu{\bf Reliable\\\bf auto. eval.}}
\\
 & & & {E} & {N} & {C} & &
\\\midrule
Free text & e-SNLI &  
\lvlyes & \lvlyes & \lvlyes & \lvlyes & \lvlmaybe & \lvlno \\
Word highlights or span-based rel. & e-SNLI 
& \lvlyes & \lvlyes & \lvlyes & \lvlyes & \lvlno & \lvlyes \\
Trees of facts \& rules/conclusions & RuleTakers 
& \lvlno & \lvlyes & \lvlno & \lvlyes & \lvlmaybe & \lvlyes \\
Trees of facts \& conclusions & EntailmentBank  
& \lvlno & \lvlyes & \lvlno & \lvlno & \lvlmaybe & \lvlyes \\
Free text with chain-of-thought & any 
& \lvlyes & \lvlyes & \lvlyes & \lvlyes & \lvlyes & \lvlno \\
Semantic trees (proposed idea) & NA 
& \lvlyes & \lvlyes & \lvlmaybe & \lvlyes & \lvlyes & \lvlyes \\
\bottomrule
\end{tabular}
}
\caption{
Characterization of explainable NLI tasks/approaches and their corresponding datasets.
\lvlmaybe{} is a middle value between \lvlyes{} (yes) and \lvlno{} (no).  
We consider \textbf{naturalness}, \textbf{comprehensiveness of explanations}, and \textbf{reliable automatic evaluation} as the most crucial features for reliably and thoroughly measuring the reasoning skills of current state-of-the-art reasoning systems, which are based on PLMs.  
}
\label{tab:compare_datasets}
\end{table*}
%████████████

%███████████████████████████████████████████████████
\section{State of the art in explainable natural language reasoning}
\label{sec:shortcomings}

The section characterizes several notable initiatives for explainable reasoning.
While doing so, we emphasize three properties: the naturalness of sentences and reasoning patterns, the comprehensiveness of explanations, and the reliability of the automatic evaluation method.
For the reliable, comprehensive, and challenging benchmarking for explainable natural reasoning, we regard these properties as crucial. 
The characteristics are summarized in \autoref{tab:compare_datasets}. 

%███████████████
\paragraph{Free-text explanations in e-SNLI} 

We believe that creating the e-SNLI dataset \citep{esnli:18} was the first significant step towards explainable NLI.
It is obtained by augmenting the SNLI dataset \citep{bowman-EtAl:2015:EMNLP} with human-elicited explanations in a natural language form using crowdsourcing: one explanation was collected for training samples and three for development and evaluation samples.

While e-SNLI provides ample training data (570K problems) for deep neural networks (DNNs), unfortunately, its automated evaluation method for explanations is not reliable.
In particular, \citet{esnli:18} concluded this by showing that mediocre system-produced explanations and the third reference explanations obtained a BLEU score similar to the first two reference explanations.
A somewhat similar finding was concluded by \citet{wang-etal-2020-semeval}: for the crowdsourced explanations, a BLEU score did not correlate well with human evaluation, where explanations provided a reason why nonsensical sentences do not make sense.

Due to the high language variation in the explanations of e-SNLI,%
\footnote{
We also found out that on average the longest explanations ($\approx$18.3 tokens) in the development part of e-SNLI are twice as long as the shortest explanations ($\approx$9.1 tokens).   
}
when using e-SNLI for evaluation, the papers randomly draw 50-100 gold explanations and manually compare them to the corresponding system-produced explanations. 
Such a way of evaluation is time-consuming, not representative, and difficult to replicate.
While e-SNLI is a large information-rich dataset, unfortunately, it fails to serve as a reliable benchmark for explainable NLI.

%███████████████
\paragraph{Word highlights as explanations in e-SNLI} 

For completeness, we would like to mention that \mbox{e-SNLI} also comes with highlighted words that crowd workers consider crucial for explaining inference relations.
The highlighted words are a side product of eliciting explanations from crowd workers as they had to use these words in their free-text explanations.
While highlighted words are a part of the explanation of inferences, they cannot be regarded as explanations.
For instance, consider an NLI problem in \eqref{ex:rte_problem}, taken from the development set of RTE-2 \citep{bar-haim-etal-2006-second}.
The highlighted spans are crucial to the explanation of the gold entailment label.
However, they are missing other necessary information such as the semantic relations between the spans. 
\bgroup
% \setlength{\abovedisplayskip}{4pt}
% \setlength{\belowdisplayskip}{4pt}
% \setlength{\abovedisplayshortskip}{0pt}
% \setlength{\belowdisplayshortskip}{0pt}
%\openup-2\jot
\begin{align}
& 
\begin{tabular}{@{}c@{:~} p{62mm}@{}}
P & The drugs that \hl{slow down} or \hl{halt} Alzheimer's disease work best the earlier you administer them\\\hline
H & Alzheimer’s disease is \hl{treated} using drugs
\hfill \textsc{entailment}
\end{tabular}
\label{ex:rte_problem}
\end{align}
\egroup

Several recent works \citep{stacey-etal-2022-logical,wu2023weakly,choudhury-etal-2023-explaining} employ semantic relations over text spans as explanations, e.g., \emph{slow down} entailing \emph{treat}, and \emph{halt} entailing \emph{treat}.
Even with this knowledge, substantial information is lacking to form a comprehensive explanation of the entailment relation for \eqref{ex:rte_problem}.
From the perspective of pre-trained Language Models (PLMs), this knowledge is not sufficient to attribute sound reasoning to a PLM-based NLI system as the knowledge might have implicitly existed in the PLM. 
While span-based semantic relations are a prerequisite for inference explanations and suitable for reliable automatic evaluation (e.g., using F1 score for matching relations),
they do not represent fine-grained explanations.

%███████████████
\paragraph{Fact-rule proofs} 

Building upon the RuleTaker's datasets \citep{ruletakers:2020}, \citet{saha-etal-2020-prover} and \citet{tafjord-etal-2021-proofwriter} prepared corresponding datasets for explainable NLI.%
\footnote{
While there are several datasets differing in details such as open- and close-world assumptions, or rules or conclusions as intermediate nodes, we focus on the general description of the task.  
}
In a nutshell, a sample of the dataset represents a pair of a premise text and a hypothesis sentence, where the premise text is a collection of sentences representing facts (e.g., \textit{Erin is blue}) and rules (e.g., \textit{If someone is blue then they are kind}).
Based on the fact and rule premises, a system is expected to predict whether a hypothesis sentence is true or false and provide a corresponding proof tree.
The proof tree is rooted in the hypothesis sentence ID, other nodes represent fact and rule IDs, and the child-parent relation stands for a rule application to the facts.
The proofs serve as explanations, and they are suitable for automatic evaluation due to their unambiguous structure.
During the evaluation, a system-produced proof is correct if it fully matches the reference proof.

Formatting explanations in terms of proof trees on facts and rules is a step forward from the evaluation perspective.
On the downside, the facts and rules in the datasets are synthetically created, which results in unnatural sentences with little diversity in syntactic structure and semantic phenomena.
Note that the diversity in syntactic structures was alleviated in ParaRules--the version of the dataset, where facts and rules were paraphrased with crowdsourcing. 
However, the reasoning patterns remain synthetic and lack the coverage of various semantic phenomena.  
Some versions of the dataset (with the closed-world assumption) include only entailment and contradiction relations. 
The version with the open-world assumption allows the neutral relation but no proof is defined for it.%
\footnote{
Providing explanations for neutral relations is challenging as there is no proof similar to entailment and contradiction relations.
Somewhat related, the fact verification task \citep{thorne-etal-2018-fact} also omits an explanation for \textsc{NotEnoughInfo} label in contrast to the \textsc{Supported} and \textsc{Refuted} labels.  
}
While facts and rules are mostly of a synthetic structure, strictly speaking, it is not explained how, for example, a rule is applied to a fact, e.g., does a model somehow match the antecedent of the rule to the fact, or is the rule application accidental?
For example, it is well known that humans tend to have difficulty with using the modus tollens rule \citep{Rips:1994,Wason:68}.
Hence, it will be preferable that an inference system, in addition to inferring the conclusion from the fact and the rule (see \eqref{ex:ruletakesr_problem} from \cite{ruletakers:2020}), provides evidence that an adequate inference rule was used for the inference.
\bgroup
% \setlength{\abovedisplayskip}{4pt}
% \setlength{\belowdisplayskip}{4pt}
% \setlength{\abovedisplayshortskip}{0pt}
% \setlength{\belowdisplayshortskip}{0pt}
%\openup-2\jot
\begin{align}
& 
\kern-2mm\begin{tabular}{@{}c@{~} p{60mm}@{}}
Fact: & The cat likes the rabbit\\
Rule: & If someone likes the bald eagle then they do not like the rabbit\\\hline
Conc: & The cat does not like the bald eagle\\
  & \hfill \textsc{entailment}
\end{tabular}
\label{ex:ruletakesr_problem}
\end{align}
\egroup
%
%

%███████████████
\paragraph{Entailment Bank} 

Yet another version of explainable NLI with structured explanations was introduced by \citet{dalvi-etal-2021-explaining}.
They created the Entailment Bank which consists of up to 2K entailment trees.
The entailment tree is structurally similar to the fact-rule proofs but differs from them in the following: 
(a) The sentences are more naturalistic and diverse, collected from the grade-school science questions;
(b) trees are manually created by trained annotators;
(c) some non-terminal nodes in the tree are not among the input facts and need to be generated on the fly.  

Although the entailment trees improve upon the fact-rule proofs in terms of the naturalness of the sentences and inferences, they have several shortcomings:
(i) The Entailment Bank focuses entirely on the entailment relation while the contradiction and neutral relations are skipped;
(ii) Creating entailment trees is a time-consuming process ($\approx$20 minutes per tree as reported by \citet{dalvi-etal-2021-explaining});
(iii) Similarly to the fact-rule proofs, entailment trees do not explain how sentences entail another sentence.
In other words, each step in the entailment trees can be seen as an instance of a simple NLI problem labeled as entailment while no explanation is provided for such sub-problems;
(iv) The evaluation based on the tree matching is too strict as it penalizes any paraphrases of the intermediate nodes generated on the fly. 

%███████████████
\paragraph{Chain-of-thought method}

Applying the method of Chain-of-thought \cite{wei2022chain,kojima2022large} to NLI problems might be seen as an approach to explainable reasoning.
In this approach, a PLM produces a free-text explanation for a predicted inference label.
However, it is not the case that whenever the chain-of-thought method predicts a correct label, the explanation is automatically correct.
The low results of the experiments by \cite{ye2022the} on e-SNLI confirm this.
Note that the chain-of-thought method suffers from the same shortcomings as any method producing free-text explanations---the absence of reliable automatic evaluation methods.

\bigskip
It seems the variability in free-text explanations is not sufficiently constrained by an NLI problem, unlike how reference translations are constrained by the source text in machine translation. 
That is why the comparison metrics commonly used in machine translation have not successfully detected the semantic equivalence of explanations in NLI.
To support an accurate automatic evaluation of explanations, one needs to abandon free-text explanations and opt for structured explanations.
However, existing approaches to generating structured explanations come with shortcomings.
If the generation is automatized, it results in synthetic data, while the manual construction of explanations is hard and time-consuming.
In the next section, we propose a reasoning framework that can model explanations for a wide range of inferences and yet has a perspective of automatization for scale-up.

%████████████
% \bgroup
% \setlength{\textfloatsep}{0pt}
\begin{figure*}[t!]
\centering
\scalebox{.92}{\begin{forest}
for tree={align=center, parent anchor=south, child anchor=north, l sep=7mm, s sep=10mm}
[\lab{1}~\natformT{The drugs that slow down or halt Alzheimer's disease}\\
\natformT{work best the earlier you administer them}\\
\mlform{S1 : $\T$}
\\\toppad{5mm}
{\lab{2}}~\natformF{Alzheimer's disease is treated using drugs}\\
\mlform{S2 : $\F$}
,baseline
  [\lab{3}~\natformT{drugs that slow down or halt Alzheimer's disease} : \args{$d$}\\
  \mlform{S1[4:52] : $d$ : $\T$}
  ,labelA={\rulecol{$\exists$}\ndList{1}}
    [\lab{4}~\natformT{drugs} : \args{$d$}\\
    \mlform{S1[4:9] : $d$ : $\T$}
    \\\toppad{5mm}
    \lab{5}~\natformT{slow down or halt Alzheimer's disease} : \args{$d$}\\
    \mlform{S1[15:52] : $d$ : $\T$}
    ,labelA={\rulecol{$\wedge$}\ndList{3}}
    ,labB={3pt}{\footnotesize}{$\vee$~\ndList{5}}
      [\lab{6}~\natformT{$d$ slow down Alzheimer's disease}\\
      \mlform{$d$\tokglue{}S1[15:24]\tokglue{}S1[33:52] : $\T$}
        [\lab{8}~\natformF{Alzheimer's disease is treated using $d$}\\
        \mlform{S2[0:36]\tokglue{}$d$ : $\F$}
        ,labelA={\rulecol{\texttt{substitute}}\ndList{2,4}}
          [$\btimes$
          ,labelA={\rulecol{\texttt{frame\_alt}}\ndList{6,8}}
          ]
        ]
      ]
      [\lab{7}~\natformT{$d$ halt Alzheimer's disease}\\
      \mlform{$d$\tokglue{}S1[28:52] : $\T$}
        [\lab{9}~\natformF{Alzheimer's disease is treated using $d$}\\
        \mlform{S2[0:36]\tokglue{}$d$ : $\F$}
        ,labelA={\rulecol{\texttt{substitute}}\ndList{2,4}}
          [$\btimes$
          ,labelA={\rulecol{\texttt{frame\_alt}}\ndList{7,9}}
          ]
        ]
      ]
    ]
  ]
]
\end{forest}}
% \vspace{-3mm}
\caption{A tableau-based proof for the NLI problem in \eqref{ex:rte_problem}.
The nodes with green and red backgrounds are natural language representations and their role is to assist the reader in understanding the machine-readable representations (in blue font color), which are formatted in terms of character offsets.
The green and red colors stand for the true and false signs, which are explicated in the machine-readable representation.
Each branching is annotated with a rule application that is responsible for the expansion of the tree.
Each rule application consists of a rule name and the ID(s) of entries to which the rule was applied. 
The tableau represents a proof for the entailment of \eqref{ex:rte_problem} as both branches are closed due to inconsistencies in the branches, representing a failure to refute the entailment relation.  
}
\label{fig:drugs_treat}
\end{figure*}
% \egroup
%████████████

%███████████████████████████████████████████████████
\section{Refutation-based reasoning framework}
\label{sec:framework}

We propose a reasoning framework that is based on the semantic tableau \cite{beth:55,Hintikka:55}, a proof procedure that has been adapted to several formal logics.
The semantic tableau is assumed to be ``one of the most popular,
since it appears to bring together the proof-theoretical and the semantical
approaches to the presentation of a logical system and is also very intuitive.''%
\footnote{From the preface of \cite{tableauhandbook:99} by Melvin Fitting.
}

The core idea of the framework is that it proves inference relations via refutation---if it fails to refute a certain relation, then it is proved.
The tableau in \autoref{fig:drugs_treat} represents a proof for the entailment of the NLI problem in \eqref{ex:rte_problem}.
The proof starts to refute the entailment by searching for a counterexample, where the premise is true and the hypothesis is false.
\lab{3} is a result of decomposing the meaning of \lab{1}, where an entity $d$ is introduced satisfying the property expressed by the noun phrase in \lab{3}.
The meaning of \lab{3} is further decomposed in terms of \lab{4} and \lab{5} by treating the relative pronoun \emph{that} as conjunction.
\lab{6} and \lab{7} represent two possible scenarios that are triggered by the disjunction in \lab{5}.
In both branches, we can apply the same rule to obtain identical entries \lab{8} and \lab{9}.
They are a result of \lab{2}, which says \emph{no drugs treat AD}, and \lab{4}, which says that $d$ is a drug.
Both \lab{8} and \lab{9} say that it is false that AD is treated using $d$.
The left branch has a contradiction as \lab{6} and \lab{8} assert contradictory information.
The same case is on the right branch.
All branches closed means that no counterexample was found for the entailment relation, hence, the entailment relation is proved.

Our reasoning framework also comes close to the mental models of \cite{johnsonlaird83} studies in cognitive psychology.
According to \cite{Johnson-Laird1991-JOHD-9}, in the first stage, a reasoner forms an implicit model based on a premise and verifies whether a putative conclusion holds in the model.
If the conclusion holds in the model, a prudent reasoner will go to the second stage to verify whether any model of the premise makes the conclusion true.
In other words, the prudent reasoner will search for a counterexample for the conclusion, i.e., a model where the premise is true and the conclusion false.

%███████████████████████████████████████████████████
\section{Designing new explainable NLI tasks}
\label{sec:tasks}

In this section, we describe how necessary data can be semi-automatically obtained and how several explainable NLI tasks with structured explanations can be defined based on the collected data.  

% \enlargethispage{3mm}

%█████████████████████████████████████
\subsection{Natural language theorem prover}
\label{ssec:nltp}

The state-of-the-art performance on mainstream NLI datasets is undoubtedly held by DNN-based NLI models.
Despite that, there have been active development of a couple of NLI systems based on formal logic and theorem proving.
The merits of logic-based NLI systems are the following: 
their precision is considerably high due to the underlying methodology, 
and their decision procedure is inherently interpretable.
To the best of our knowledge, only two logic-based systems have managed to reach the human-performance level (84\%) on SICK \citep{marelli-etal-2014-sick}:
% \footnote{
% The SICK dataset has relatively simple and short sentences and targets lexically rich and compositional inferences.
% The dataset poses challenges for logic-based systems in obtaining logical forms of relatively wide-coverage constructions and usage of lexical knowledge.   
% % Usually, the FraCaS dataset \citep{fracas} represents an entry point for such systems.
% }
\textsc{ccg2lambda} \citep{mineshima-etal-2015-higher,yanaka-etal-2018-acquisition} and \textsc{LangPro} \citep{abzianidze-2015-tableau,abzianidze-2017-langpro}. 
From these two systems, we find \textsc{LangPro} to be a better fit for collecting data for the new explainable NLI tasks for three reasons: 
(i) It has a higher precision on SICK (94\% vs 84\%);
(ii) its underlying natural logic uses logical forms that are not very specific to a certain meaning representation (in contrast to first-order logic);
(iii) the structure of its proofs is faithful to the underlying proof calculus, the semantic tableau method \citep{beth:55,Hintikka:55}. 

%████████████
% \bgroup
% \setlength{\textfloatsep}{0pt}
\begin{figure}[t!]
\centering
\scalebox{.92}{\begin{forest}
for tree={align=center, parent anchor=south, child anchor=north, l sep=7mm, s sep=10mm}
[{\lab{1}~$\sysm{many}_{\nou,\vp,\sen}~ \sysm{bird}_\nou (\sysm{high}_{\vp,\vp}~ \sysm{hover}_\vp) : \elist : \T$}\\
\surfform{many birds hover high}
\\\toppad{6mm}
{\lab{2}~$\sysm{few}_{\nou,\vp,\sen}~ \sysm{bird}_\nou~ \sysm{fly}_\vp : \elist : \T$}\\
\surfform{few birds fly}
, baseline
,labB={3pt}{\footnotesize}{\mbox{\upDisCov}~\ndList{1,2}}
 [{\lab{3}~
 $\sysm{high}~ \sysm{hover} : [c] : \T$}\\
 \surfform{$c$ hover high}
 \\\toppad{6mm}
  {\lab{4}~
 $\sysm{fly} : [c] : \F$}\\
 \surfform{$c$ fly}
  [{\lab{7}~
   $\sysm{hover} : [c] : \T$}\\
   \surfform{$c$ hover}
   ,labelA={\rulen{adj$^\subset_\T$}\ndList{3}}
   [{\lab{8}~
    $\btimes$}
    ,labelA={\clrule\ndList{4,7}}
   ]
  ]
 ]
 [{\lab{5}~
 $\sysm{many}~ \sysm{bird} : [\sysm{fly}] : \T$}\\
 \surfform{many birds fly}
 \\\toppad{6mm}
  {\lab{6}~
 $\sysm{few}~ \sysm{bird} : [\sysm{fly}] : \T$}\\
 \surfform{few birds fly}
  [{\lab{9}~
  $\sysm{many} : [\sysm{bird}, \sysm{fly}] : \T$}\\
  \surfform{many birds fly}
  ,labelA={\rulen{a\push}\ndList{5}}
   [{\lab{10}~
   $\sysm{few} : [\sysm{bird}, \sysm{fly}] : \T$}\\
   \surfform{few birds fly}
   ,labelA={\rulen{a\push}\ndList{6}}
    [{\lab{11}~
    $\btimes$}
    ,labelA={$\btimes\mid$~\ndList{9,10}}
    ]
   ]
  ]
 ]
] 
\end{forest}}
% \vspace{-3mm}
\caption{The tableau proves that \quot{many birds hover high} contradicts \quot{few birds fly} and vice versa.
}
\label{fig:tableau_proof}
\end{figure}
% \egroup
%████████████

The proofs produced by \textsc{LangPro} will be the source of the data for the proposed tasks.
\autoref{fig:tableau_proof} illustrates a tableau proof from \textsc{LangPro}.
The nodes of the trees represent terms of simply typed $\lambda$-calculus backed up with the semantics of higher-order logic \citep{church:1940}.  
The main idea behind the proof is to find a counterexample for the target relation, e.g., the contradiction relation.
A counterexample of contradiction is a situation where the premise and the hypothesis are true, hence the tableau proof starts with \lab{1,2} marked with $\T$.
The tableau grows by applying rules from the predefined set of rules to the existing nodes and breaking the antecedent terms into smaller pieces.
The annotations of the edges show this process.
For instance, \lab{3-6} nodes are obtained by applying (\upDisCov) rule to \lab{1,2}.%
\footnote{
The rule (\upDisCov) exploits upward monotonicity of \sysm{many} in its second argument, e.g., \emph{many birds $V_1$} entails \emph{many birds $V_2$} if $V_1$ entails $V_2$.
The idea behind the rule is as follows. 
If $f^{\uparrow}(x)=1$ and $g(y)=1$ where $f$ and $g$ are Boolean functions and $f$ is upward monotone, then there are two possibilities: either $x \not< y$ (the left branch) or $x < y$, hence $f^{\uparrow}(y)=1$ (the right branch).
Note that $A~B~C : []$, $A~B : [C]$, and $A : [B,C]$ are semantically the same nodes only differing in terms of formatting that distinguishes functions from their argument list.
% More details about the rule can be found in \cite{abzianidzethesis}.
}
The branching represents two possibilities in searching for a counterexample.
In the example, both branches are closed due to inconsistencies spotted in the corresponding possible situations.
The left branch represents an inconsistent situation due to some entity $c$ hovering but not flying (see \lab{4,7}) while the right branch is closed as \sysm{few} and \sysm{many} cannot be true on the same arguments.
The tableau with all branches closed represents a failure to find a counterexample.
Therefore, the relation for which the refutation failed is proved and the closed tableau represents its proof.

%█████████████████████████████████████
\subsection{From proofs to explainable NLI tasks}
\label{ssec:tasks}

% \url{https://github.com/kovvalsky/LangPro}

%█████████████████████████
\subsubsection{Answering several concerns}
\label{sssec:qa}

Before describing the design of specific tasks, we would like to answer several questions that naturally arise about the methodology that utilizes proofs as explanations.

%████████████
\paragraph{Is it feasible to get a substantial amount of proofs?}

\textsc{LangPro} achieves $84\%$ accuracy on SICK with $94\%$ of precision \citep{abzianidze-2020-learning}, however, many false proofs are due to noisy gold labels in SICK.
While SICK is not a challenging dataset, similar NLI problems do occur in other NLI datasets. 
Considering the number and the size of the existing human-annotated NLI datasets (in total more than 1M problems), it is realistic to obtain proofs for at least 20K of NLI problems.%
\footnote{
We expect most of these problems to be drawn from SICK, SNLI, MED \citep{yanaka-etal-2020-neural}, Breaking NLI \citep{glockner-etal-2018-breaking}, NegNLI \citep{hossain-etal-2020-analysis}, and ConjNLI \citep{saha-etal-2020-conjnli}. 
}
This is already an order of magnitude larger than the Entailment Bank.
The advantage of the method is that the data collection process is automatized and produces detailed proofs with extremely high consistency and correctness.
For instance, matching such levels of detail, consistency, and correctness will also be challenging with trained annotators.

% \enlargethispage{3mm}
%████████████
\paragraph{Will not the proved NLI problems be homogeneous?}

The number of inference rules used to build tableau proofs is up to 80.
The rules cover syntactic categories and constructions such as prepositional phrases, expletives, light verb constructions, open compounds, verb subcategorization, passives, copula, and coordination.
This will allow sufficient syntactic diversity in the proved problems and will represent a substantial improvement over the existing inference datasets with structured explanations \citep{saha-etal-2020-prover,tafjord-etal-2021-proofwriter,dalvi-etal-2021-explaining}.
In the collected proofs, there might be relatively many proofs that are structurally identical to other proofs.
However, such proofs will differ from each other in terms of lexical entities and pose a challenge for systems from a generalization perspective.   

%████████████
\paragraph{But human reasoning is different from deductive reasoning}

The inferences that humans validate but \textsc{LangPro} fails to prove will be out of reach when collecting proofs with the help of \textsc{LangPro}.
The NLI problems that \textsc{LangPro} proves but get different inference labels from human annotators will be manually verified by experts.   
In many cases, the gold label can be overridden by \textsc{LangPro}'s prediction given that human annotations often disagree with each other for various reasons. 
This means that some of the problems that were complex for crowd workers to correctly classify will be identified by \textsc{LangPro} and be eventually correctly classified.
We see it unnecessary to dumb down NLI problems by aligning their labels to crowd workers' judgments.
For example, it is known that humans are not good at using \textit{modus tollens} rule \cite{Wason:68}, but this does not mean that we need to eliminate the use of \textit{modus tollens} when reasoning with NLI problems.

%█████████████████████████
\subsubsection{New explainable NLI tasks}
\label{sssec:tasks}

We describe a list of explainable NLI tasks that will be defined by the datasets derived from the tableau proofs of \textsc{LangPro}.
We present the list following the order based on the explanation richness.  
An illustration sample of the datasets will be based on the proof in \autoref{fig:tableau_proof}.

%████████████
\paragraph{Lexical relations as explanations}

One of the simplest structured explanations is a set of lexical relations that are necessary for the inference.
For example, the explanation of the NLI problem in \autoref{fig:tableau_proof} could be two lexical relations: $\sysm{hover}\sqsubset\sysm{fly}$ and $\sysm{many}|\sysm{few}$, where the relations stand for the subsumption and disjoint/alternation relations.
All such necessary relations can be easily extracted from tableau proofs.
In some cases, the relations can be over the short phrases if it is not further decomposable, e.g., $\sysm{mouse}\sqsubset\sysm{small animal}$. 
The NLI problems that use no lexical relation, e.g., \emph{Not all birds fly} entailing \emph{Some bid does not fly}, will simply carry an empty explanation.
Optionally, such problems can also be omitted for this type of explanation.

%████████████
\paragraph{Rules and lexical relations as explanations}

To provide a relatively simple explanation for NLI problems that require no lexical relations, we propose to augment the (possibly empty) set of lexical relations with the multiset of inference rules used in the tableau proof.
In the case of the example proof, the multiset of rules will be [(\upDisCov),(\rulen{adj$^\subset_\T$})].
Here, we ignore the argument pushing rule (\rulen{a\push}) as it has no semantic contribution and also omit ($\btimes\mid$) and (\clrule) as they are redundant beside the lexical relations.
The multiset of the example entailment requiring no lexical relation will be [($\neg$), (\allF), (\someF), ($\neg$)].  
The multiset reflects the complexity of the entailment: two negation rules correspond to \emph{not} and \emph{does not} while the rules for universal and existential quantifiers correspond to \emph{all} and \emph{some} where their subscript makes sure that \emph{all} is in the negative polarity context and \emph{some} in the hypothesis.  

In the current and the previous tasks, we recommend using F-score.
In the case of multisets of inference rules, the comparison should be based on exact matching to avoid inflating the score.

%████████████
\paragraph{Unlabeled proofs as explanations}

To add more details and structure to the explanations, we propose the proof trees without inference rule labels as explanations in addition to the lexical relations.
The example of an unlabeled proof tree is illustrated in \autoref{fig:tableau_proof_compact}.
Alternatively, the lexical relations can be encoded in terms of the closure rules and their references to the antecedent nodes.

It will be too much to ask an NLI model to generate exact $\lambda$-terms of the nodes as this automatically includes semantic parsing tasks when considering the initial nodes.
Given that we want to prevent NLI models from committing to a certain meaning representation, the node entries in the gold proofs will be converted into the original surface forms (as shown with gray text in \autoref{fig:tableau_proof}).
Fortunately, this is possible with the help of the naturalness of the $\lambda$-terms, as each term represents a constituent or a constituent with a trace.
It is important to emphasize the naturalness of natural logic formulas.
With the help of this property, NLI systems are not forced to learn semantic parsing along with reasoning.
Moreover, to decrease the depth of the trees, it is recommended to omit the structure-related inference rules, e.g., (\rulen{a\push}), that have no semantic contribution. 
This simplification results in a shorter right-hand side branch in \autoref{fig:tableau_proof_compact}. 

One might find some visual resemblance between the unlabeled proof trees and the entailment trees of the Entailment Bank \citep{dalvi-etal-2021-explaining}.
However, this resemblance is superficial.
In entailment trees, the tree edges represent entailment arrows while in tableau proofs the edges are conjunctions;
A branch in a tableau represents a situation modeling a conjunction of all terms sitting on the branch.

% is  the former has several advantages over the latter:
% (i) In addition to entailment, the unlabeled proof trees model the contradiction and neutral relations;
% (ii) They carry out reasoning till the lexical level;
% (iii) Collecting unlabeled proof trees can be automatized.

%████████████
% \bgroup
% \setlength{\textfloatsep}{0pt}
\begin{figure}[t!]
\centering
\scalebox{.92}{\begin{forest}
for tree={align=center, parent anchor=south, child anchor=north, l sep=7mm, s sep=7mm}
[{$\texttt{many birds hover high} : \T$}
\\\toppad{6mm}
{$\texttt{few birds fly} : \T$}
, baseline
% ,labB={3pt}{\footnotesize}{\mbox{\upDisCov}~\ndList{1,2}}
 [{$\texttt{$c$ hover high} : \T$}
 \\\toppad{6mm}
  {$\texttt{$c$ fly} : \F$}
  [{$\texttt{$c$ hover} : \T$}\\
   $\btimes$
   % ,labelA={\rulen{adj$^\subset_\T$}\ndList{3}}
  ]
 ]
 [{$\texttt{many birds fly} : \T$}
 \\\toppad{6mm}
  {$\texttt{few birds fly} : \T$}\\
 $\btimes$
 ]
] 
\end{forest}}
% \vspace{-3mm}
\caption{The unlabeled proof represents the simplified version of the proof from \autoref{fig:tableau_proof}.
Note that all word forms in the nodes are identical to those in the premise and the hypothesis. 
}
\label{fig:tableau_proof_compact}
\end{figure}
% \egroup
%████████████

%████████████
\paragraph{Complete proofs as explanations}

The final version of the explainable NLI task employs the entire proof as an explanation.
The lexical relations and the inference rules will be all encoded in the tree.
Like in the unlabeled proof trees, the trees here will also be shortened by pruning the semantically vacuous inference rules.

% \enlargethispage{3mm}
%███████████████████████████████████████████████████
\section{Conclusion}

We have presented a reasoning framework that accounts for diverse semantic phenomena and its explanations (i.e.) proofs are suitable for reliable automatic evaluation.
Searching for counterexamples is a key function of the framework.
It is motivated by the semantic table method, a popular proof system for formal logics.
It also shared similarities with the psychology of reasoning, in particular, the theory of mental models: 
``cornerstone of human rationality---insofar as humans are rational---is that
they grasp the force of counterexamples'' \citep{Johnson-Laird2006-JOHHWR}.

We have presented four explainable NLI tasks that exploit tableau proofs to extract relevant information that serves as structured explanations.
The key components that make this feasible are three characteristics of the \textsc{LangPro}: the naturalness of terms, an intuitively simple reasoning calculus, and almost perfect precision of the prover with a decent accuracy.

All the described tasks are compatible with the automatic evaluation as their structured explanations do not suffer from linguistic variability: each lexical item can be represented in terms of character offsets in an NLI problem. 
The granularity of the explanations becomes finer when moving from the simplest task to the most information-rich one.
In all four tasks, explanations are based on fine-grained units such as phrases and lexical units.
The tasks cover both entailment and contradiction relations.
Structuring explanations for neutral cases require special care and we leave this for the feature research.

We have started building a proof bank based on the described approach.
The proof bank consists of NLI problems and their corresponding formal proofs with structured explanations in various formats.
While the term \emph{formal proof} might suggest a long and entangled sequence of symbols, in fact, the tableau proofs for NLI problems are user-friendly as illustrated in \autoref{fig:tableau_proof_compact}.

%███████████████████████████████████████████████████
\section{Limitations}

This is a position paper and the proposed approach has some limitations.
More limitations might appear when it is put to an application.
The current limitations are the following:
(i) The proposed semantic framework is sub-optimal as it cannot accommodate all types of natural language reasoning such as abduction or inductive reasoning;
(ii) There are several semantic phenomena in deductive reasoning that are not covered by the proposed framework (e.g., including counting, intensionality, comparatives, etc.), and future research is required to gauge the extent of this limitation.
(iii) The naturalness of the sentence structures and inference patterns that are claimed and attributed to the proposed reasoning framework is relative to the existing approaches.

%███████████████████████████████████████████████████
% \section{To add}
% \begin{itemize}
% \item Emphasize that it is a framework not about a specific tool
% \item We are not using a tool to solve problems but to generate data. This is a common practice of generating artificial data, but the tool we use is much better.
% \item Natural logic is suitable for LLMs as they are good on surface forms. They do some reasoning but the fragment of that reasoning is simple (hit current success papers).
% \item Give 2-3 examples of tableau proofs of non-trivial examples.
% \item Learning tree is not a problem, as it has been shown by entailment trees that it works.
% \item Hit entailment trees for intermediate issues.
% \item Make a table that lists issues and does comparison across approaches?
% \end{itemize}

% Acknowledgement Johan Bos

\bibliography{mine}
\bibliographystyle{acl_natbib}

% \appendix

% \section{Example Appendix}
% \label{sec:appendix}

% This is a section in the appendix.

\end{document}